\documentclass{article}      

\usepackage{amsmath} 
\usepackage{amssymb}  
\usepackage{algorithmicx}
\usepackage{algpseudocode}
\usepackage{xcolor}
\usepackage{todonotes}
\usepackage{subcaption}
\usepackage{booktabs}
\usepackage{multirow}
\usepackage{url}
\usepackage{nips15submit_e,times}
\usepackage{siunitx}

\usepackage[english]{babel}
\usepackage[square, numbers]{natbib}
\DeclareMathOperator{\softmax}{softmax}

\DeclareMathOperator{\vis}{vis}
\DeclareMathOperator{\VL}{VL}

\DeclareMathOperator{\sz}{size}
\DeclareMathOperator{\sign}{sign}
\DeclareMathOperator*{\argmax}{argmax}
\DeclareMathOperator*{\argmin}{argmin}

\newcommand{\mb}[1]{\mathbf{#1}}
\newcommand{\vx}{\mathbf{x}}
\newcommand{\pd}[2]{\frac{\partial}{\partial #2} #1}
\newcommand{\otoprule}{\midrule[\heavyrulewidth]}
\newcommand{\vl}[1]{\VL_{#1}}

\algnewcommand{\LineComment}[1]{\State \(\triangleright\) #1}

\nipsfinalcopy

\begin{document}

\title{
Visualization Regularizers for Neural Network based Image Recognition
}

\author{Biswajit Paria\\
CSE, IIT Kharagpur, India\\
biswajitsc@iitkgp.ac.in
\And Vikas Reddy\\
Mining Engg., IIT Kharagpur, India\\
vikas.challaram@iitkgp.ac.in
\And Anirban Santara\\
CSE, IIT Kharagpur, India\\
anirban\_santara@iitkgp.ac.in
\And Pabitra Mitra\\
CSE, IIT Kharagpur, India\\
pabitra@cse.iitkgp.ernet.in
}


\maketitle

\begin{abstract}

The success of deep neural networks is mostly due their ability to learn meaningful features from the data. Features learned in the hidden layers of deep neural networks trained in computer vision tasks have been shown to be similar to mid-level vision features. We leverage this fact in this work and propose the visualization regularizer for image tasks. The proposed regularization technique enforces smoothness of the features learned by hidden nodes and turns out to be a special case of Tikhonov regularization. We achieve higher classification accuracy as compared to existing regularizers such as the L2 norm regularizer and dropout, on benchmark datasets without changing the training computational complexity.

\end{abstract}

\section{Introduction}
Regularization is an important aspect of deep neural network training, to prevent over-fitting in the absence of sufficient data. Usually regularizers restrict the norms of the weight parameters. A commonly used class of regularizers, the Lp norm regularizers, penalize the Lp norms of the weight parameters. Of these, the L1 and L2 norm regularizers are most popular. Other regularizers include soft-weight sharing \cite{nowlan1992simplifying}, layer-wise unsupervised pre-training \cite{bengio2007greedy}, and dropout \cite{srivastava2014dropout}. It has been shown by \citet{erhan2010does} that layer-wise unsupervised pre-training has a regularizing effect during training. There has also been recent works on adaptive dropout \cite{ba2013adaptive}, which is an improvement over the original dropout. In this paper, we propose a novel regularizer which we call as the visualization regularizer (VR), based on the visual quality of features learned by the hidden nodes. We introduce two variants of the VR regularizer, based on the $\|\cdot\|_1$ and $\|\cdot\|_2$ norms respectively.

Vision tasks benefit from features consisting of primitives recognized by mid-level vision systems. Deep neural networks have been known to learn hierarchical layers of feature representation \cite{lecun2015deep, bengio2009learning}. On observing the features learned by deep neural networks trained using back propagation, it is seen that in contrast to well defined mid-level features, the node features are often noisy. More meaningful features (smoother features for example) can be favorable to training the network. Our proposed regularizer imposes a constraint on the hidden nodes of a neural network to learn smoother features.  Since, the definition of the regularizer depends on the visual property of smoothness, it is only pertinent to domains with a notion of spatial locality, such as images.

We show that the VR regularizer is a special case of Tikhonov regularization \cite{tikhonov1977solutions}. The Tikhonov matrix of the conventional L2 norm regularizer corresponds to an identity matrix multiplied by the regularization weight. Whereas, the Tikhonov matrix for the VR regularizer is more generalised but sparse.

We perform experiments with our regularizer on two benchmark datasets: MNIST \cite{lecun1998mnist} and CIFAR-10 \cite{Krizhevsky2009}. We observe that our regularizer aids in learning and improves the classification accuracy when used alongside other regularizers. However the computational complexity of the VR regularized training algorithm remains the same as that of the unregularized training algorithm.

This paper is organized in five sections. Section II gives a brief introduction to the architecture of deep neural networks and the notation used. The notion of visualization of a node is formally defined in section III. The proposed VR regularizer and the training algorithm along with the complexity analysis are described in section IV. Section V establishes the relationship to Tikhonov regularization. Section VI presents all the experimental results and observations. Finally Section VII concludes with a summary of the achievements and scopes for future work.

\section{Deep neural network architecture}
The investigations presented in this paper are based on a multi-class classification setting. The notation followed in this paper is as follows: $\vx$ denotes the input, $W_i$ and $\mb{b}_i$ respectively denote the weights and biases corresponding of the layers, $\mb{u}_i$ denotes the pre-activation of the layers, $\mb{h}_i$ denotes the activation of layers and $\mb{y}$ denotes the output of the neural network. The subscripts $i$ in the notation denotes the $i$th hidden layer. The following equations describe a deep neural network with $l$ hidden layers.
\begin{eqnarray}
  \mb{u}_1 &=& W_1\cdot\vx+\mb{b}_1,\\
  \label{eqn:g} \mb{h}_i &=& g(\mb{u}_i),\ \forall 1\le i \le l,\\
  \mb{u}_{i+1} &=& W_{i+1}\cdot \mb{h}_i + \mb{b}_{i+1},\ \forall 1\le i \le l-1,\\
  \mb{y} &=& \softmax(W_{l+1}\cdot \mb{h}_l + \mb{b}_{l+1}),
\end{eqnarray}
where $g$, is the activation function, a monotonically increasing non-linear function such as sigmoid, tanh or the rectified linear unit~\cite{glorot2011deep}.

The loss function used for training is the sum of the classification loss and the regularization term. For a neural network $M$ and a dataset $D$, the loss function can be written as,
\begin{equation}
L(M, D) = L_c(M,D) + \lambda Reg(M),
\end{equation}
where $L_c(M,D)$ denotes the classification loss between the output of the neural network and the true class labels, $Reg(M)$ denotes the regularization term and $\lambda$ denotes the regularizer weight. However, dropout cannot be included in the loss function. It is incorporated into the training algorithm.

\section{The notion of visualization of a node}

Visualization of a node refers to the visualization of the features learned by the node. Visualization of the nodes of a neural network has been studied by \citet{erhan2009visualizing}. They proposed the activation maximization algorithm for visualizing features learned by a node. Following the notion in \cite{erhan2009visualizing} we define the visualization of a node as follows.

Visualization of a node is defined to be the input pattern(s) which activate(s) the node maximally under the restriction of the L2 norm of the input to be equal to unity. The L2 norm of the input is restricted to unity to prevent the input from becoming unbounded.
Formally, the visualization $\vis(n)$ of a node $n$ is defined as,
\begin{equation}
  \label{eqn:visdef}
  \vis(n) = \displaystyle \argmax\limits_{\|\vx\|_2 = 1} n(\vx),
\end{equation}
where $n(\vx)$ denotes the activation of node $n$ for input $\vx$. Depending on the non-linearity $g$ used in equation \ref{eqn:g}, solution to the above equation can be unique or multiple. For example, there exists an unique solution for invertible functions like $g = \sigma$ (sigmoid) or $g = \tanh$. Whereas there can be multiple solutions for non-invertible functions such as $g(\cdot) = \max(0, \cdot)$ also known as the rectified linear unit (ReLU). The visualization of an internal node of the neural network can be computed using gradient-ascent as described by \citet{erhan2009visualizing} as activation maximization.

Observe that the pre-activation of a node $n$ in the \textit{first} hidden layer for an input vector $\vx$ is $\mb{w^\intercal}\vx$, where $\mb{w}$ denotes the weights of the connections coming into the node $n$. The pre-activation is maximized when $\vx$ is aligned in the same direction as $\mb{w}$ in the appropriate vector space. Since the activation function is a monotonically increasing function, maximization of the pre-activation maximizes the activation too. Consequently, we get the following closed form as one of the possible solutions to equation \ref{eqn:visdef} for nodes $n$ in the first hidden layer.
\begin{equation}
  \label{eqn:visexp}
  \vis(n) = \frac{\mb{w}}{\|\mb{w}\|_2}
\end{equation}
Note that finding a closed form algebraic expression for the visualization of nodes in the higher hidden layers is difficult due to the non-linearity of the activation function.


\section{Proposed visualization based regularizer}

The VR regularizer is based on the expression of visualization as given in equation~\ref{eqn:visexp}. We utilize equation~\ref{eqn:visexp} and produce an appropriate regularization loss which we include in final training loss. The following subsections give a detailed description of the VR regularizer.

\subsection{Smoothness of a visualization}
Intuitively, one can determine whether an image is smooth or noisy by looking at the gradients in the image. An image is smooth if it has small gradients. The gradient of an image can be computed by convolving it with a 2D high pass filter. Examples of high pass filters include, first order gradient filters such as the Sobel operator, or second order gradient filters such as the Laplacian operator. Larger the pixel values in the convolution, the larger the gradients in the original image, and greater presence of noise in the image. We utilize this intuition to give a formal definition of smoothness.

Consider a convolution $I\otimes K$ of image $I$ with kernel $K$, where $K$ is a high pass filter like the laplacian kernel. We define the smoothness of an image to be the negative sum of squares of pixel values of $I\otimes K$. Equivalently, we can define the visualization loss $\VL$ of image $I$ as,

\begin{eqnarray}
  \nonumber I' &=& I \otimes K,\\
  \VL(I) &=& \sum_{i,j} (I'\cdot I')_{ij},
  \label{eqn:VLSqSum}
\end{eqnarray}

where ``$\cdot$'' denotes the element-wise product, also known as the Schur or Hadamard product. The visualization loss is the negative of the smoothness of an image. Lower the visualization loss, smoother is the image. Table \ref{tab:LossValues} shows the visualization loss for some example visualizations.

\begin{table}
\centering
\caption{Visualization loss of example visualizations.}
\label{tab:LossValues}
\begin{tabular}{ccc}
\toprule Image ($I$)& Convolution ($I \otimes K$) & $\VL(I)$ \\
\otoprule    \begin{minipage}{0.1\textwidth}
		\centering
	    \includegraphics[width=0.6\textwidth]{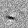}
          \end{minipage}
          &
          \begin{minipage}{0.1\textwidth}
	    \centering
	    \vspace{0.1cm}
	    \includegraphics[width=0.6\textwidth]{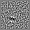}
	    \vspace{0.1cm}
          \end{minipage}
          & 135.6717\\
\midrule    \begin{minipage}{0.1\textwidth}
	    \centering
	    \includegraphics[width=0.6\textwidth]{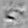}
          \end{minipage}
          &
          \begin{minipage}{0.1\textwidth}
	    \centering
	    \vspace{0.1cm}
	    \includegraphics[width=0.6\textwidth]{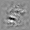}
	    \vspace{0.1cm}
          \end{minipage}
          & 16.1213\\
\midrule    \begin{minipage}{0.1\textwidth}
	    \centering
	    \includegraphics[width=0.6\textwidth]{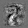}
          \end{minipage}
          &
          \begin{minipage}{0.1\textwidth}
	    \centering
	    \vspace{0.1cm}
	    \includegraphics[width=0.6\textwidth]{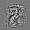}
	    \vspace{0.1cm}
          \end{minipage}
          & 9.3421\\
\midrule    \begin{minipage}{0.1\textwidth}
	    \centering
	    \includegraphics[width=0.6\textwidth]{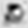}
          \end{minipage}
          &
          \begin{minipage}{0.1\textwidth}
	    \centering
	    \vspace{0.1cm}
	    \includegraphics[width=0.6\textwidth]{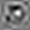}
	    \vspace{0.1cm}
          \end{minipage}
          & 0.0685\\
\bottomrule
\end{tabular}
\end{table}

\subsection{Visualization loss as a regularizer}

Classification tasks using deep neural networks benefit from the high-level of abstractions achieved in the higher layers of the neural network. Deep neural networks are intended to utilize low-level pixels to learn mid-level features and finally high-level features. We propose the visualization regularizer (VR) to constrain the nodes in the first hidden layer to learn features with qualities similar to mid-level visual features. This constraint is intended to facilitate the discovery of high-level abstractions more effectively.

Informally, we define the VR regularizer as a regularizer to reduce the visualization loss of the nodes of the neural network. In other words the VR regularizer makes the nodes learn smooth or less noisy features.

The following sub-sections give a more detailed description of the VR regularizer.
\subsubsection{Regularizer expression}
Let $U(M)$ denote the set of nodes of the first hidden layer of a neural network $M$ and for a node $n \in U(M)$, let $\mb{w}_n$ denote the weights of the connections incoming into the node.

From equation \ref{eqn:visexp} we know that visualization of a node in the first hidden layer is proportional to the weights of the connections coming into the node. Hence the visualization loss of the node is proportional to the visualization loss of the weight vector coming into the node. Therefore we can use $\VL(\mb{w}_n)$ as a surrogate for the visualization loss of the visualization of the node $n$. The difficulty of computing an algebraic expression for the visualization of nodes in higher hidden layers, limits the usage of the surrogate to nodes in the first hidden layer only.

We define the visualization loss $\VL$ of a neural network $M$ as,

\begin{equation}
 \VL(M) = \sum_{n \in U(M)} \VL(\mb{w}_n).
 \label{eqn:RegExpM}
\end{equation}

The network training loss function can thus be defined as,
\begin{equation}
 L(M,D) = L_c(M,D) + \mu \VL(M) + \lambda L_2'(M),
 \label{eqn:LossFunction}
\end{equation}
where $L_2'(M)$ denotes the L2 norm regularization term for all weights except the weights coming into the first hidden layer. The remaining notation are as described in section 2.

\subsubsection{Gradient of the visualization regularizer}
For the visualization loss to be used as a regularizer, its gradient must be computed with respect to the model parameters. Automatic gradient computation libraries such as Tensorflow~\cite{tensorflow} and Theano~\cite{theano}, obviate the need to compute the gradients manually. However they are an additional overhead on computational resources, and hence for manual computation of gradients for increased efficiency, we derive the expression of the derivative of the VR regularizer. This is also useful in the case of embedded computing, where computational resources are limited. In the following paragraphs we derive the gradient for a general kernel $K$ of size $(2k+1) \times (2k+1)$.

For simplicity in computing the expression of the gradient, we index the elements of the kernel relative to the central element as shown in equation \ref{eqn:KIndex}. The element at the center is indexed $(0,0)$. All other elements are indexed according to their position relative to the central element.
\begin{equation}
  K =
  \begin{bmatrix}
    a_{-k,-k} & \hdots & a_{-k,0} & \hdots & a_{-k,k} \\
    \vdots   &  \ddots & \vdots   & \ddots & \vdots\\
    a_{0,-k} & \hdots & a_{0,0} & \hdots & a_{0,k} \\
    \vdots   &  \ddots & \vdots   & \ddots & \vdots\\
    a_{k,-k} & \hdots & a_{k,0} & \hdots & a_{k,k}
  \end{bmatrix}
  \label{eqn:KIndex}
\end{equation}

Let $N_K$ denote the set of indices of the kernel matrix.
\begin{equation}
  N_K=\{(i,j)\ |\ i, j \in \{-k, \dots, k\} \}
\end{equation}
  
Consider an image $I$ with dimensions $n, m$. Let $S(i,j)$, corresponding to the $(i,j)$th pixel of $I$, be defined as follows.
\begin{eqnarray}
  \nonumber S(i,j) &=& \{(r, (p,q))\ |\ (p,q) = (i,j) + r,\ r \in N_K \\
   & & \mbox{ and } 0 \le p < n,\ 0\le q < m\}
\end{eqnarray}
Informally, $S(i,j)$ contains the set of valid indices $(p,q)$, along with their position $r$ relative to $(i,j)$, that need to be considered while computing the convolution for the $(i,j)$th pixel.

A full convolution of the image $I$ can be described as
\begin{equation}
  \label{eqn:convexp}
  I' = I \otimes K = \left[ \sum_{(r, (p,q)) \in S(i,j)}a_r I_{pq}\right]_{ij}.
\end{equation}
It follows from the definition that a pixel $I_{pq}$ present at a position $r$, relative to $(i,j)$, has the coefficient $a_r$ in $I'_{ij}$.

The visualization loss is
\begin{equation}
  \VL(I) = \sum_{i,j} {I'}_{ij}^2.
  \label{eqn:VLSum}
\end{equation}

The partial derivative of the visualization loss with respect to a pixel $I_{ij}$ of the image $I$ is
\begin{equation}
  \pd{\VL(I)}{I_{ij}} = \sum_{p,q} \pd{{I'}_{pq}^2}{I_{ij}}.
\end{equation}
  
Observe that in the above equation $I_{ij}$ occurs in $I'_{pq}$ only for $(p,q)$ such that $(r, (p,q)) \in S(i,j)$. Moreover, if $(p,q)$ is present at position $r$ relative to $(i,j)$, then $(i,j)$ is present at position $-r$ relative to $(p,q)$. It follows that $I_{ij}$ has the coefficient $a_{-r}$ in $I'_{pq}$. Hence the derivative can be computed as,
\begin{eqnarray*}
  \pd{\VL(I)}{I_{ij}} &=& \sum_{(r, (p,q)) \in S(i,j)} \pd{{I'}_{pq}^2}{I_{ij}}\\
  &=& \sum_{(r,(p,q)) \in S(i,j)} 2I'_{pq} \pd{I'_{pq}}{I_{ij}}\\
  &=& \sum_{(r,(p,q)) \in S(i,j)} 2I'_{pq} a_{-r}.\\
\end{eqnarray*}

Further, using equation~\ref{eqn:convexp}, we can write, 
\begin{equation}
  \label{eqn:visgrad}
  \pd{\VL(I)}{I} = 2 (I'\otimes K^I) = 2 ((I \otimes K) \otimes K^I),
\end{equation}
where $K^I$ denotes the kernel matrix formed by flipping $K$ both horizontally and vertically. Formally,
\begin{equation}
K^I_{ij} = K_{(-i,-j)}.
\end{equation}

\begin{figure}
\centering
\includegraphics[width=0.7\textwidth]{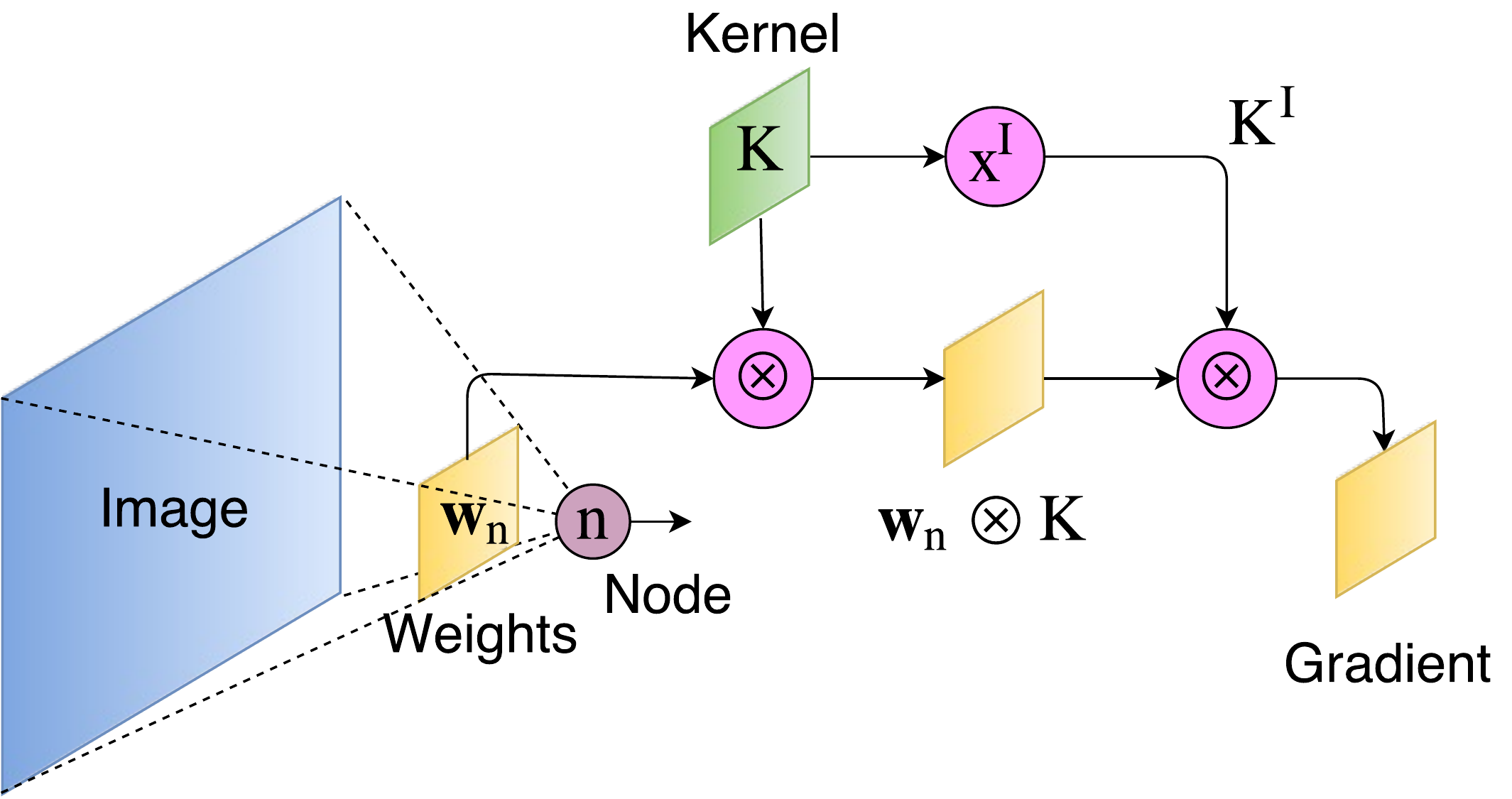}
\caption{Computation of gradient for visualization loss. $\otimes$ denotes convolution.}
\label{fig:VLossDiag}
\end{figure}

Equation \ref{eqn:visgrad} allows us to compute the gradient efficiently and in a scalable manner. All popular GPU programming frameworks provide libraries for scalable convolutions. Figure \ref{fig:VLossDiag} illustrates the gradient computation for the visualization loss.

\subsubsection{Regularized training algorithm}

Training requires computing the gradient for the regularized loss function with respect to the parameters of the network.
As evident from equation \ref{eqn:LossFunction}, the gradient of the loss function can be computed by first computing the gradients of $L_c(M,D)$, $L_2'(M)$ and $\VL(M)$, and then computing their sum. The gradients of $L_c(M, D)$ and $L_2'(M)$ can be computed using back-propagation and partial derivatives respectively.

By equation \ref{eqn:RegExpM}, the gradient of $\VL(M)$ is the sum of gradients of $\VL(\mb{w}_n)$ for $n \in U(M)$. Note that the gradient $\pd{\VL(\mb{w}_n)}{w}$ is zero if $w \notin \mb{w}_n$. In other words, the gradient is zero if $w$ does not belong to the set of weights incoming to node $n$. Hence we only need to compute the gradients $\pd{\VL(\mb{w}_n)}{\mb{w}_n}$ for all $n \in U(M)$. These can be computed using equation \ref{eqn:visgrad}. The full algorithm described in figure~\ref{fig:trainingalg}. The algorithm can be extended to using dropout and momentum.

\begin{figure}
 \begin{algorithmic}[1]
 \Procedure{Train}{$D,M,K,\mu,\lambda,\alpha$}
  \LineComment{$D$: the training data, $M$: the model respectively.}
  \LineComment{$K$: Kernel for the VR regularizer.}
  \LineComment{$\mu$: VR regularizer weight, $\lambda$: L2 regularizer weight.}
  \LineComment{$\alpha$: learning rate.}
  \LineComment{Returns trained model $M$}
  \State
  \LineComment{Get the weight variables from the model}
  \State $\mb{w} := w(M)$
  \State Initialize $\mb{w}$
  \State
  \While{does not converge}
    \For{minibatch sample $D_m$}
      \LineComment{Step 1: Compute using back-propagation.}
      \State $\mb{u} \gets \nabla_\mb{w} L_c(D_m, M)$
      \LineComment{Step 2: Gradient for L2 regularizer.}
      \State $\mb{v} \gets \nabla_\mb{w} L_2'(M)$ 
      \State
      \LineComment{Step 3: Gradient for VR regularizer.}
      \LineComment{Iterate over the nodes in the 1st fully connected layer.}
      \For{$n \in U(M)$} 
	\LineComment{Compute $\mb{z}_n = \pd{\VL(\mb{w}_n)}{\mb{w}_n}$}
	\LineComment{using equation \ref{eqn:visgrad}}
    \State $\mb{z}_n = 2 ((\mb{w}_n \otimes K) \otimes K^I)$
      \EndFor
      \LineComment{Concatenate the gradients.}
      \State $\mb{z} = (\dots,\mb{z}_n,\dots)$
      \State
      \LineComment{Step 4: Compute total gradient.}
      \State $\mb{g} \gets \mb{u} + \lambda\mb{v} + \mu\mb{z}$
      \LineComment{Step 5: Take step.}
      \State $\mb{w} \gets \mb{w} - \alpha \mb{g}$
    \EndFor
  \EndWhile
  \State return $M$
 \EndProcedure
 \end{algorithmic}
 \caption{Regularized training algorithm}
 \label{fig:trainingalg}
\end{figure}

Computing each $\mb{z}_n$ in figure \ref{fig:trainingalg} takes $O(|\mb{w}_n|)$. Consequently, computing the gradient of the VR regularizer takes $O(|\mb{w}|)$ time, which is of the same order as computing the gradient of the L2 norm regularizer. Thus, the VR regularizer does not impose additional overhead in the computational complexity per iteration.

\subsection{A variant of the VR regularizer}
In our description of the VR regularizer we have defined the smoothness of an image $I$ as the sum of squares of the convoluted image $I'$ (equation~\ref{eqn:VLSqSum}), which is actually the square of the 2-norm ($\|\cdot\|_2$) of the flattened vector of $I'$. Analogously, we can also define an 1-norm variant of the VR regularizer. To distinguish between the 1-norm and 2-norm variants, we denote the respective losses by $\vl{1}$ and $\vl{2}$. Similar to the definition of $\vl{2}$ in equation~\ref{eqn:VLSqSum}, $\vl{1}$ can be defined as follows.
\begin{eqnarray}
  \nonumber I' &=& I \otimes K,\\
  \VL_1(I) &=& \sum_{i,j} |I'_{ij}|.
  \label{eqn:VL1SqSum}
\end{eqnarray}

Thus, the training loss can be modified as,
\begin{equation}
 L(M,D) = L_c(M,D) + \mu_1 \VL_1(M) + \mu_2 \VL_2(M) + \lambda L_2'(M),
 \label{eqn:Loss1Function}
\end{equation}

Similar to the expression of the gradient for $VL_2$ (equation~\ref{eqn:visgrad}), we can derive the gradient of $VL_1$ as

\begin{equation}
  \label{eqn:vis1grad}
  \pd{\VL_1(I)}{I} = 2 (\sign(I')\otimes K^I) = 2 (\sign(I \otimes K) \otimes K^I),
\end{equation}

where $\sign(M)$ denotes the elementwise application of the signum\footnote{$\sign(x)=\begin{cases} 1, &\text{if }x > 0\\ 0, &\text{if }x=0\\ -1, &\text{if } x < 0\end{cases}$} function on the elements of the matrix $M$.

\section{Relationship with Tikhonov regularization}

Tikhonov regularization was originally developed for solutions to ill-posed problems \cite{tikhonov1977solutions}. For example, L2 regularization, a special case of Tikhonov regularization is used to compute solutions to regression problems for which rank deficient matrices are encountered while computing their solutions. The solution to regularized least squares regression is given by,

\begin{equation}
 w = \argmin_w \| X w - y \|^2 + \lambda \| \Gamma w \|^2,
\end{equation}

where $\Gamma $ is the Tikhonov matrix. The L2 regularizer corresponds to $\Gamma = I$.

We show that the $\vl{2}$ regularizer is also a special case of Tikhonov regularization. Let $\mb{w}$ be the concatenation of all the weights in $\mb{w}_n$ for $n \in U(M)$. From equation \ref{eqn:VLSum}, we can see that $\VL(\mb{w}_n)$ is the sum of squared terms of the form ${\mb{w}'_{ij}}^2 = (\sum_t c_t w_t)^2$, where $w_t \in \mb{w}$, $c_t \in K$ and $\mb{w}' = \mb{w}_n \otimes K$. For example $ \mb{w}'_{ij} = (w_p-w_q-w_r)^2$ can be represented in the given form where $c_p = 1, c_q = -1, c_r = -1$, and all other $c_t$ are zero.

The Tikhonov matrix $\Gamma$ can be constructed as follows. Consider the expression of $\VL(M) = \sum _{n \in U(M)} \VL(\mb{w}_n)$ consisting of the sum of $p$ such squared terms. Let the $i$th term in this expression be $s_i = (\sum_t c_{it} w_t)^2$. Then, $\Gamma$ and consequently $\VL(M)$ in terms of $\Gamma$ are, respectively,

\begin{equation}
\Gamma = \left[c_{it}\right], \mbox{ and } \VL(M) = \| \Gamma \mathbf{w} \|^2.
\end{equation}
In practice, it can be assumed that the kernel matrix $K$ is has a constant size. Hence it follows, that the expression of $s_i$ consists of only a constant number of non-zero $c_{it}$ since the number of non-zero $c_{it}$ is bounded by the size of the kernel matrix $K$. Hence, the number of non-zero entries in $\Gamma $ is $O(p\cdot\sz(K)) = O(p)$, whereas the total number of entries in $\Gamma$ is $ p|\mb{w}|$, concluding the sparsity of $\Gamma$.

\section{Experiments and observations}
We experimented on the MNIST \cite{lecun1998mnist} and CIFAR-10 \cite{Krizhevsky2009} datasets and compared the classification accuracy of our algorithm using the VR regularizers with other regularizers.\footnote{The experiment code in Theano is available at \url{https://github.com/biswajitsc/VisRegDL}. An implementation in Tensorflow is also available at \url{https://github.com/cvikasreddy/VisReg}.}

\subsection{Experimental setting}
For classification of MNIST digits, we experimented with fully connected architectures and convolutional architectures, and we only experimented with convolutional architectures for the classification of CIFAR-10 objects. For fully connected architectures, we applied the VR regularizer to the weights of layer immediately after the input layer i.e. the first hidden layer. For convolutions however, we applied the VR regularizer to the weights of the fully connected layer immediately after the last convolutional layer. The detailed layerwise descriptions of the architectures used in our experiments are given in table~\ref{tab:arch_description}.

We used the laplacian kernel for the VR regularizer defined as follows.
\begin{equation}
 K = \begin{bmatrix}
      -1 & -1 & -1 \\
      -1 & 8 & -1 \\
      -1 & -1 & -1 \\
     \end{bmatrix}
\end{equation}

We used the mean cross-entropy loss over the mini-batches as the classification loss. The total training loss is as defined in equation~\ref{eqn:Loss1Function}. We trained the neural network using stochastic gradient descent with momentum \cite{nesterov1983method}.

The the initial learning rate for all the models was fixed to 0.01. The model was trained for 2300 epochs and 250 epochs for CIFAR-10 and MNIST respectively. For CIFAR-10, the learning rate was reduced by a factor of 1.3 every 500 epochs. For MNIST, it was reduced to 0.005 after the 75th epoch and reduced by a factor of 1.3 thereafter every 25th epoch. For finding the optimal values of the regularization weights, we performed a randomized hyper-parameter search with manual fine-tuning.

\begin{table}
\centering
\caption{Description of the network architectures used in our experiments. fc($n$) denotes a fully connected layer with $n$ nodes, dropout($p$) denotes a dropout layer with nodes dropped with probability $p$, conv($s\times s$, $c$) denotes a convolutional layer with a kernel of size $s\times s$ and $c$ output channels, and maxpool($s \times s$) denotes a maxpooling layer with a $s \times s$ sized kernel.}
\label{tab:arch_description}
\begin{tabular}{lc}
\toprule
Dataset / Architecture & Description \\\otoprule
MNIST / fully connected & \begin{minipage}{0.6\textwidth}
                           \centering
                           input(784) -- fc(1000) -- dropout(0.3) -- fc(1000) -- dropout(0.3) -- fc(1000) -- output(10)
                          \end{minipage}\\\midrule
MNIST / convolutional   & \begin{minipage}{0.6\textwidth}
                           \centering
                           input(28$\times$28) -- conv(3$\times$3, 64) -- conv(3$\times$3, 64) -- dropout(0.1) -- maxpool(3$\times$3) -- dropout(0.1) -- fc(1024) -- output(10)
                          \end{minipage}\\\midrule
CIFAR-10 / convolutional& \begin{minipage}{0.6\textwidth}
                           \centering
                           input(32$\times$32) -- conv(5$\times$5, 64) -- dropout(0.1) -- maxpool(3$\times$3) -- conv(5$\times$5, 64) -- dropout(0.1) -- maxpool(3$\times$3) -- conv(5$\times$5, 64) -- dropout(0.1) -- maxpool(3,3) -- dropout(0.1) -- fc(384) -- dropout(0.1) -- fc(192) -- dropout(0.1) -- out(10)
                          \end{minipage}\\\midrule
\end{tabular}
\end{table}

\begin{table}
\centering
\caption{Accuracies for various regularizer settings. `-' denotes that the corresponding regularizer was not used. $\vl{1}$ and $\vl{2}$ denote the respective mean losses over the whole dataset.}
\label{tab:results}
\begin{tabular}{lcccc}
\toprule
Dataset / Architecture & $\mu_1$ & $\mu_2$ & $\lambda$ & Acc. \% \\\otoprule
\multirow{5}*{MNIST / fully connected}
  & - 	& - 	& - 	& 98.61 \\\cmidrule{2-5}
  & - 	& - 	& 0.02 	& 98.75 \\\cmidrule{2-5}
  & - 	& 0.01 	& 0.01 	&\textbf{98.81} \\\cmidrule{2-5}
  & 0.01& -	& - 	& 98.39  \\\midrule
\multirow{5}*{MNIST / convolutional}
  & - 	& - 	& - 	& 99.21 \\\cmidrule{2-5}
  & - 	& - 	& 0.01 	& 99.16 \\\cmidrule{2-5}
  & - 	& 0.02 	& - 	& 99.27 \\\cmidrule{2-5}
  & 0.01& -	& - 	&\textbf{99.29}	\\\midrule
\multirow{4}*{CIFAR-10 / convolutional} 
  & - 	& - 	& - 	& 80.49 \\\cmidrule{2-5}
  & - 	& 0.01 	& 0.004	& 81.91 \\\cmidrule{2-5}
  & 0.01& - 	& 0.004	& 82.12 \\\cmidrule{2-5}
  & -	& 0.001	& 0.004	& 82.09 \\\cmidrule{2-5}
  &0.001& -	& 0.004	&\textbf{82.18}	\\\bottomrule
\end{tabular}
\end{table}

\subsection{Discussion}
We compared training various neural network models with combinations of L2, VR, and dropout regularizers. The accuracy and optimal hyper-parameters for various regularizer settings are given in table \ref{tab:results}. The parameters $\mu_1, \mu_2, \lambda$ and $\alpha$ respectively denote the VR regularizer weights, the L2 regularizer weight, and the learning rate.

From the table it is observed that the VR regularizer leads to an improvement in the classification accuracy. This is observed both for fully connected and convolutional neural networks. Moreover it is also observed that VR$_1$ is a better regularizer for convolutional neural networks compared to VR$_2$, whereas VR$_2$ works better in the case of fully connected neural networks.

\section{Conclusion and future work}
In this paper we introduced a new regularizer for deep neural networks trained for image tasks. We formulated the regularizer based on the notion of smoothness of a visualization, and also derived the expression of its gradient. We experimentally observe that the VR regularizer aids in learning and leads to an improvement in the classification accuracy. The VR regularizer introduces a new class of regularizers based on domain assumptions (assumption of smoothness of natural images in our case). Such kind of regularizers can also be used in other domains such as audio and video, where similar to natural images, transitions along the dimensions of the data are not noisy, but mostly smooth.

We can conclude by saying that VR regularizers are a promising direction towards more general regularization techniques which use domain knowledge.

\bibliography{ref}

\end{document}